\documentclass[runningheads]{llncs}
\usepackage[T1]{fontenc}
\usepackage{graphicx}

\usepackage{tabularx} 

\begin{document}

\title{A Fine-Grained Image Description Generation Method Based on Joint Objectives}
%
%
\author{Yifan Zhang\inst{1} \and
Chunzhen Lin\inst{1} \and
Donglin Cao\inst{1}\thanks{*Corresponding author} \and
Dazhen Lin\inst{1}}

\authorrunning{F. Author et al.}
%
\institute{Artificial Intelligence Department, Xiamen University, Xiamen 361005, China \and The Key Laboratory of Cognitive Computing and Intelligent Information Processing of  Fujian Education Institutions, Wuyi University, Wuyishan 354300, China \\
\email{another@xmu.edu.cn}
}

\maketitle              
\begin{abstract}
The goal of fine-grained image description generation techniques is to learn detailed information from images and simulate human-like descriptions that provide coherent and comprehensive textual details about the image content. Currently, most of these methods face two main challenges: description repetition and omission. Moreover, the existing evaluation metrics cannot clearly reflect the performance of models on these two issues. To address these challenges, we propose an innovative Fine-grained Image Description Generation model based on Joint Objectives. Furthermore, we introduce new object-based evaluation metrics to more intuitively assess the model's performance in handling description repetition and omission. This novel approach combines visual features at both the image level and object level to maximize their advantages and incorporates an object penalty mechanism to reduce description repetition. Experimental results demonstrate that our proposed method significantly improves the CIDEr evaluation metric, indicating its excellent performance in addressing description repetition and omission issues.
\end{abstract}
\section{Introduction}

Image description generation aims to depict visual content in natural language accurately. Presently, research in image description generation focuses on two levels: coarse-grained and fine-grained. Coarse-grained image description generation primarily addresses the image's main elements, summarizing the image content. Conversely, fine-grained image description generation aims to generate a more detailed text, delving into the image's subject, intricate details, and environmental context. Fine-grained image description generation demands a model capable of handling complex visual object relationships and generating fluent, coherent descriptions. Studying fine-grained image description generation assists in a more comprehensive understanding of image information, significantly propelling the development of image understanding and processing technology. Hence it bears significant research value.

However, due to the complex relationships between objects within images and semantic disparities between images and texts, current fine-grained image description generation methods face two main issues: (1) Repetition: Existing methods often over-emphasize the image's primary objects, resulting in repetitive and incoherent descriptions; (2) Omission: These approaches typically focus only on a few or significant objects, overlooking other objects within the image, leading to a failure to describe the image comprehensively.

Existing Methods in fine-grained image description generation \cite{chatterjee2018diverse,krause2017hierarchical,liang2017recurrent,wu2019densely,liu2018context} often applies a sequence-to-sequence construction approach, combined with hierarchical design principles. However, without hierarchical constraints, such strategies may overemphasize significant portions of the image, neglecting less obvious objects and other information, thereby failing to ensure the completeness and uniqueness of the descriptions. To address these challenges, we propose a Fine-grained Image Description Generation method based on Joint Objectives (FIDG-JO). This method establishes connections between sentences and their corresponding objects, extracts multi-object features, filters out interference from other information, and retains spatial relationship information among objects. In addition, we introduce an objective penalty mechanism, effectively mitigating issues of repetition and omission in descriptions.

Our main contributions can be summarized as follow:

\begin{itemize}
    \item We propose a novel fine-grained image description generation method based on joint objectives that establish correspondences between sentences and their respective objects. This approach extracts multi-object features corresponding to each sentence, thereby maintaining inter-object spatial relations while minimizing the interference of other information while generating descriptions. Additionally, the proposed method integrates an object penalty mechanism to reduce repetition in the descriptions.
    \item We introduce new object-based evaluation metrics that intuitively reflect the severity of repetition and omission in image description generation. By gathering statistics on the frequency of object occurrences and the total and proportional numbers of objects described, we can more accurately assess a method's performance in addressing repetition and omission issues in image description generation.
    \item Comprehensive experimental results indicate that our method significantly improves the CIDEr score, demonstrating its effectiveness in addressing description repetition and omission to a considerable extent.
\end{itemize}

\section{Related Work}

\subsection{Visual Language Models}

Current research in visual language models can be divided into two categories: understanding-based models\cite{chen2020uniter,li2020unicoder} and models based on both understanding and generation\cite{sun2019videobert}. While capable of handling a relatively limited set of tasks, the former falls short in addressing open-ended queries such as image description generation and open-ended visual question-answering tasks. The latter garners more attention owing to its broader task coverage and enhanced generalizability. Models such as SIMVLM\cite{wang2021simvlm}, Oscar\cite{li2020oscar}, OFA\cite{wang2022ofa}, and VinVL\cite{zhang2021vinvl} belong to the category of understanding and generation-based models.

Pre-training visual-language models require an object detection dataset and a paired image-text dataset. The pre-training process is complex and highly demands datasets, while data annotation is costly. Thus, to lower these dataset requirements and training costs, Wang et al.\cite{wang2021simvlm} proposed a Simplified Visual Language Model (SimVLM), which employs a prefix language model for end-to-end pre-training. SimVLM does not require pre-training for object detection or auxiliary losses, instead leveraging large-scale weak supervision to reduce training complexity. Without the need for additional data or task-specific customizations, SimVLM's pre-training approach significantly outperforms previous methods.

Traditional visual-language model training paradigms primarily adopt a pre-training finetuning approach. However, Yang et al.\cite{yang2022prompt} have integrated a new finetuning paradigm - prompt tuning\cite{brown2020language} into multimodal pre-training, significantly enhancing the model's robustness.

The research on visual-language models is evolving continuously. Despite the enduring challenges, we can anticipate the development of more effective and robust visual-language models through model improvements and the application of novel technologies.

\subsection{Fine-grained Image description generation}

Krause et al.\cite{krause2017hierarchical} were the first to introduce the task of generating paragraph-level textual descriptions for images, and they made public the Image Paragraph Captioning dataset, which is currently the primary dataset used for fine-grained image description generation. They proposed a Hierarchical Recurrent Neural Network (HRNN) for this task, where the sentence-level RNN generates a sentence topic based on image region features, and the word-level RNN generates a sentence based on the sentence topic, thus forming a complete paragraph.

In fine-grained image description generation, images often contain multiple objects. It is a challenge to determine which objects need to be described. Krause et al.\cite{krause2017hierarchical} used average pooling to encode region features into a global vector, then fed it into a Long Short-Term Memory (LSTM) network to generate the topic. However, this method may lose the inherent structural information between objects. To address this problem, Wang et al.\cite{wang2019convolutional} proposed a Convolutional Autoencoder (CAE) structure to generate image topics. The CAE abstracts and encodes the topic by performing convolution operations on region features. After that, it guides the deconvolutional decoder by reconstructing the topic to features, making the learned topics more representative.

To alleviate repetition issues and increase diversity in image descriptions, Melas et al.\cite{melas2018training} proposed a repeat penalty mechanism. They were the first to use the SCST method in fine-grained image description generation. They incorporated a trigram repeat penalty to reduce repetition by lowering the likelihood of repeated words, thereby increasing the diversity of descriptions. Additionally, Kanani et al.\cite{kanani2020improving} proposed a method that uses a language discriminator to enhance language diversity and reduce repetition by measuring word movement distance.

Existing methods for generating fine-grained image descriptions have optimized against repetitive descriptions, yet they still fall short of expectations. Moreover, there are no practical solutions to the problem of omitted descriptions. Consequently, we propose a fine-grained image description generation method based on joint objectives, aiming to mitigate both issues of repetition and omission in descriptions.

\section{Method}

To address the issues of repeated and omitted descriptions, we propose a Fine-grained Image Description Generation method based on Joint Objectives(FIDG-JO). As shown in Figure 1, FIDG-JO consists of three main components: an object feature extraction module, a combined object module, and a language module. The innovation of FIDG-JO lies in its treatment of the objects related to the current sentence through the combined object module. It extracts features from the combined objects and excludes unrelated image content. Thus, it can maintain the relationship between the objects while extracting fine-grained information from the image, minimizing interference from other image content in the current sentence.
\begin{figure}
    \centering
    \includegraphics[width=1.0\textwidth]{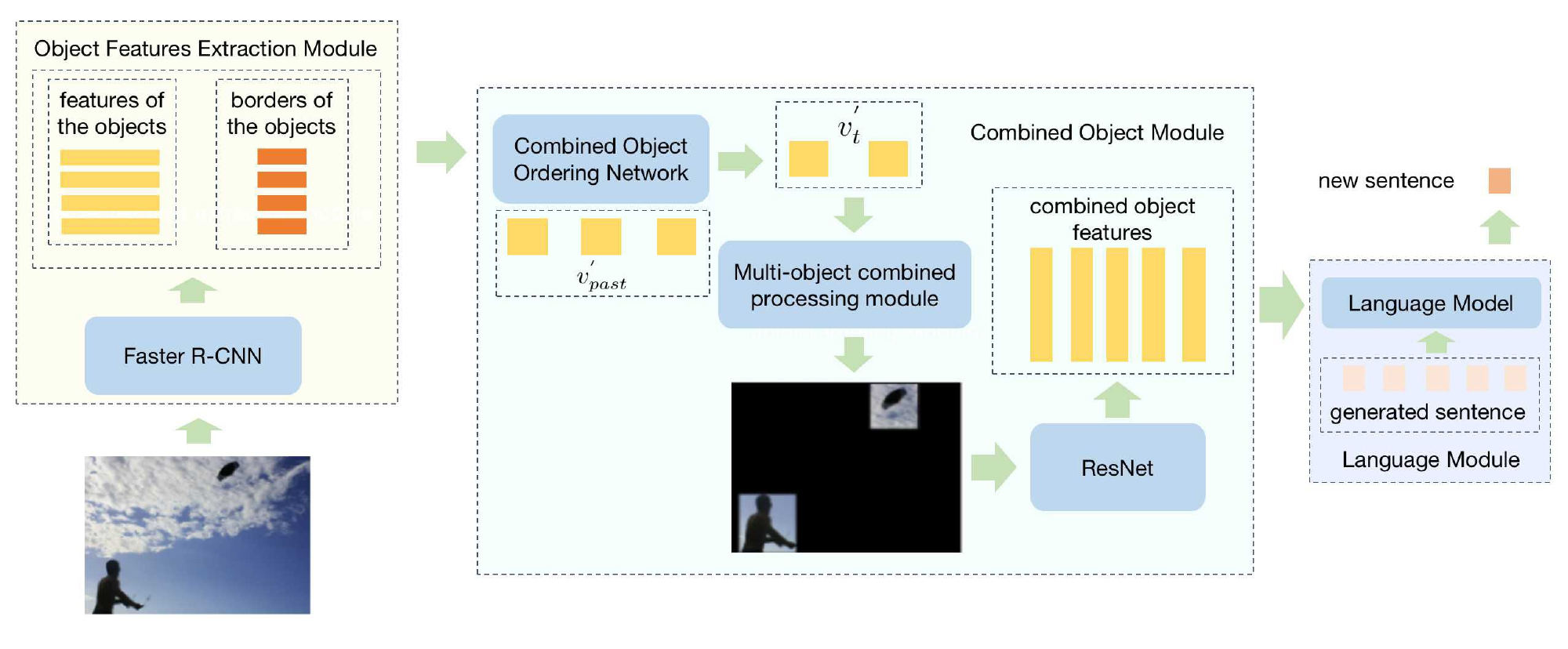}
    \caption{FIDG-JO contains three modules: an object feature extraction module, a combined object module, and a language module.}
    \label{fig:my_label}
\end{figure}

\subsection{Object Features Extraction Module}

We denote the input image as $I$ and its corresponding ground truth as $y$, comprised of $S$ sentences. The $i^{th}$ sentence consists of $N_i$ words, where $y_{i,j}$ indicates the $j^{th}$ word in the $i^{th}$ sentence.

Our object feature extraction module employs a pre-trained Faster R-CNN\cite{ren2015faster} for object detection, identifying $K$ object regions $r={r_1,r_2,...,r_K} \in R^{4*K}$ from image $I$, where $R^{4*K}$ represents a $4*K$ dimensional space. Each $r_i={B_h,B_w,B_y,B_x}$ is a 4D bounding box specifying the height ($B_h$), width ($B_w$), and top-left coordinates ($B_y,B_x$) of the object region. The associated visual features are denoted as $v={v_1,v_2,...,v_k} \in R^{H*K}$, where $R^{H*K}$ represents an $H*K$ dimensional space, with $H$ denoting the dimension of the visual features. In our experiments, we set $H=2048$. The final visual feature $v'$ for each object region is obtained by concatenating the object feature vector and the bounding box.
\begin{equation}
v' = Concat([v,r])
\end{equation}

\subsection{Combined Object Module}

"Combined objects" refers to collectively processing all $M$ objects corresponding to a sentence $y_i$, thus forming a new sub-image of combined objects to extract visual features. We introduce the combined objects module to enhance the capture of inter-object relationships and reduce the interference of other irrelevant information in generating the current sentence.

The proposed combined objects module's workflow consists of three steps: (1)object selection and ordering, (2)object combination, and (3)feature extraction.

\subsubsection{Object Selection and Ordering}

An image usually contains multiple objects, and the aim of object selection and sorting is to learn which objects need to be described, which ones should be described within the same sentence, and the order in which they should be described. We propose a combined object ordering network to select and sort the K objects in an image.

For multiple objects $O=\{O_1,O_2,...,O_K\}$ in an image, with corresponding object features $v'=\{v_1',v_2',...,v_k'\}$ and corresponding image ground truth $y=\{y_1,y_2,...,y_s\}$, the object regions relevant to the $i$-th sentence $y_i$ are $sr_i=\{sr_{i,1},sr_{i,2},...,sr_{i,Mi}\}$. We denote $SO$ as the objects that have been sorted, and $sr$ and $sv'$ represent the sorted object regions and features, respectively.

\begin{figure}
    \centering
    \includegraphics[width=1.0\textwidth]{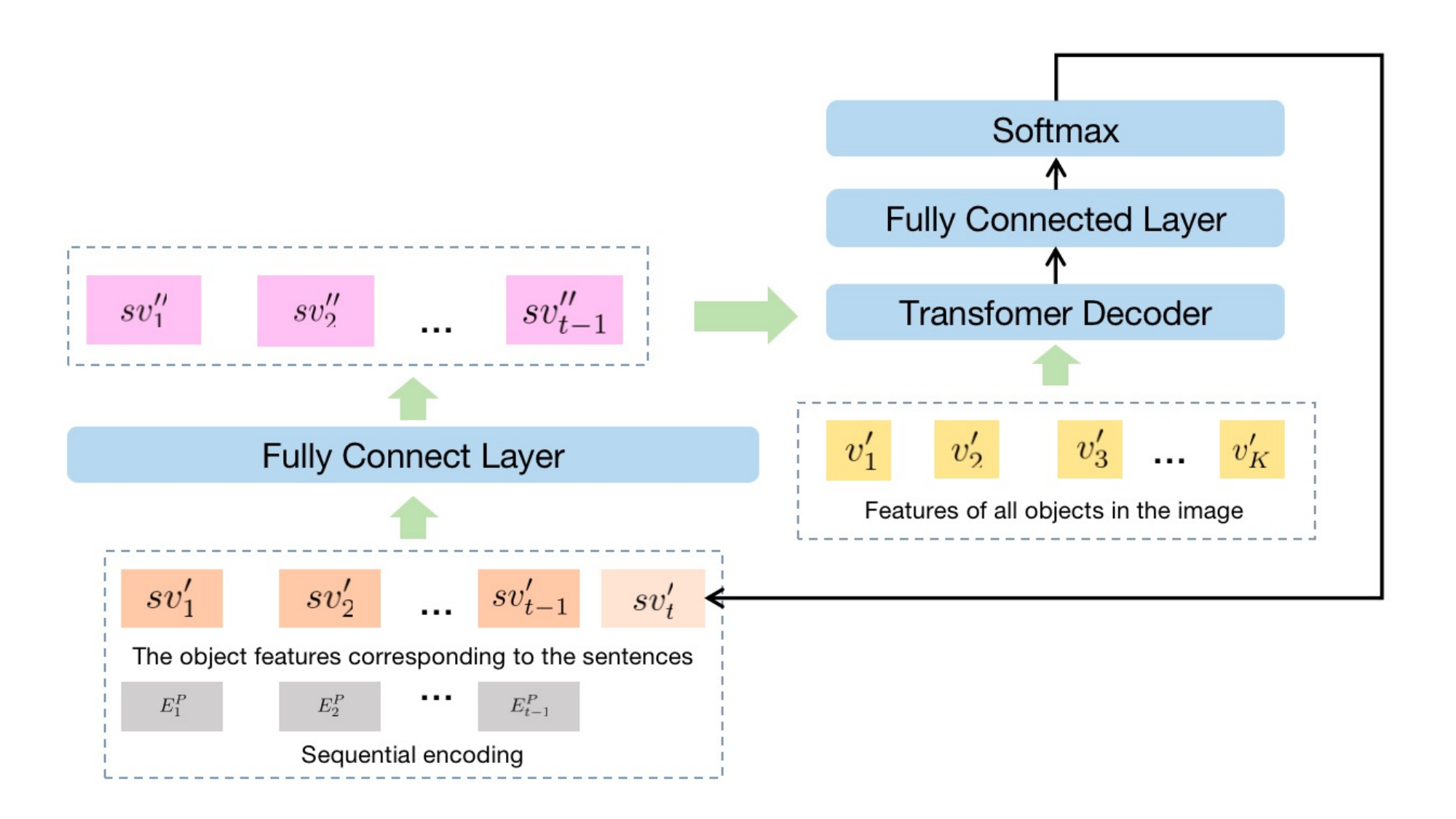}
    \caption{The framework of combined object ordering network}
    \label{fig:my_label}
\end{figure}

As illustrated in Figure 2, the structure of the combined object ordering network has two main inputs. On the left, it inputs the sequence of the combined object features $sv_{past}'=[sv_1',sv_2',...,sv_t-1']$ that have been generated for the first $t-1$ sentences, where $sv_i'$ represents the features of the object regions corresponding to the $i$-th sentence $y_i$. On the right, it inputs the object features $v'={v_1',v_2',...,v_k'}$ corresponding to the $K$ objects in the image $I$. After passing through a fully connected layer, the decoder's output is fed into a softmax layer to predict the probabilities of each object being the next object in the sequence. The calculation formula for $p$ is as follows:
\begin{equation}
    p=softmax(FC(logits))
\end{equation}
\begin{equation}
    logits=Transformer(sv_{past}''|v')
\end{equation}
where FC stands for a fully connected layer, and the left and right sections of "|" correspond to the two inputs for the Transformer decoder.
\begin{equation}
    sv''=W(sv'+E^p)+b
\end{equation}
where $E^p$ represents the positional encoding, and $W \in R^{(H+4)*(H+4)},b\in R^{H+4}$ denote learned parameters.

Additionally, to mitigate the issue of redundant object descriptions, we introduce an object penalty mechanism that reduces the probability of an object being predicted as the next object based on the number of times it has already appeared.
\begin{equation}
    logp_i=logp_i-\alpha logX_i
\end{equation}
where $p_i$ is the probability that object $O_i$ is predicted as the next object, $X_i$ is the number of times object $O_i$ has already appeared, and $\alpha$ is a hyperparameter. When $\alpha$ equals 0, the object penalty mechanism is not utilized.

\subsubsection{Object Combination}

After determining the $M$ objects through the combined object ordering network, we locate their corresponding regions in the image, retaining the information in these areas and filling other positions with zeros, forming a new combined object sub-image. This approach not only reduces interference from redundant information but also preserves the relative positional relationship between objects, enabling more effective processing of object relationships within the image. 

The steps are as follows: Compute the dimensions of all object bounding boxes within the current batch, calculating the maximum height $B_h^{max}$ and width $B_w^{max}$ among the composite objects. Then we create a batch of composite object sub-images $I^c$ with dimensions $B_h^{max}$ and $B_w^{max}$, filled with zeros, and treat each composite object as an individual image. For the M objects $\{SO_1^i,SO_2^i,...,SO_{M_i}^i\}$ corresponding to sentence $y$, we retrieve their pixel information from the original image, copy it into the combined object sub-image $I^c$, resulting in a combined object sub-image $I^c$ corresponding to the M objects of sentence $y$.

Suppose the combined objects occupy a too-small area in the image, and the background (such as the black area) has a large proportion. In that case, we will appropriately enlarge the object area to enhance its visual features.
\begin{figure}
    \centering
    \includegraphics[width=1.0\textwidth]{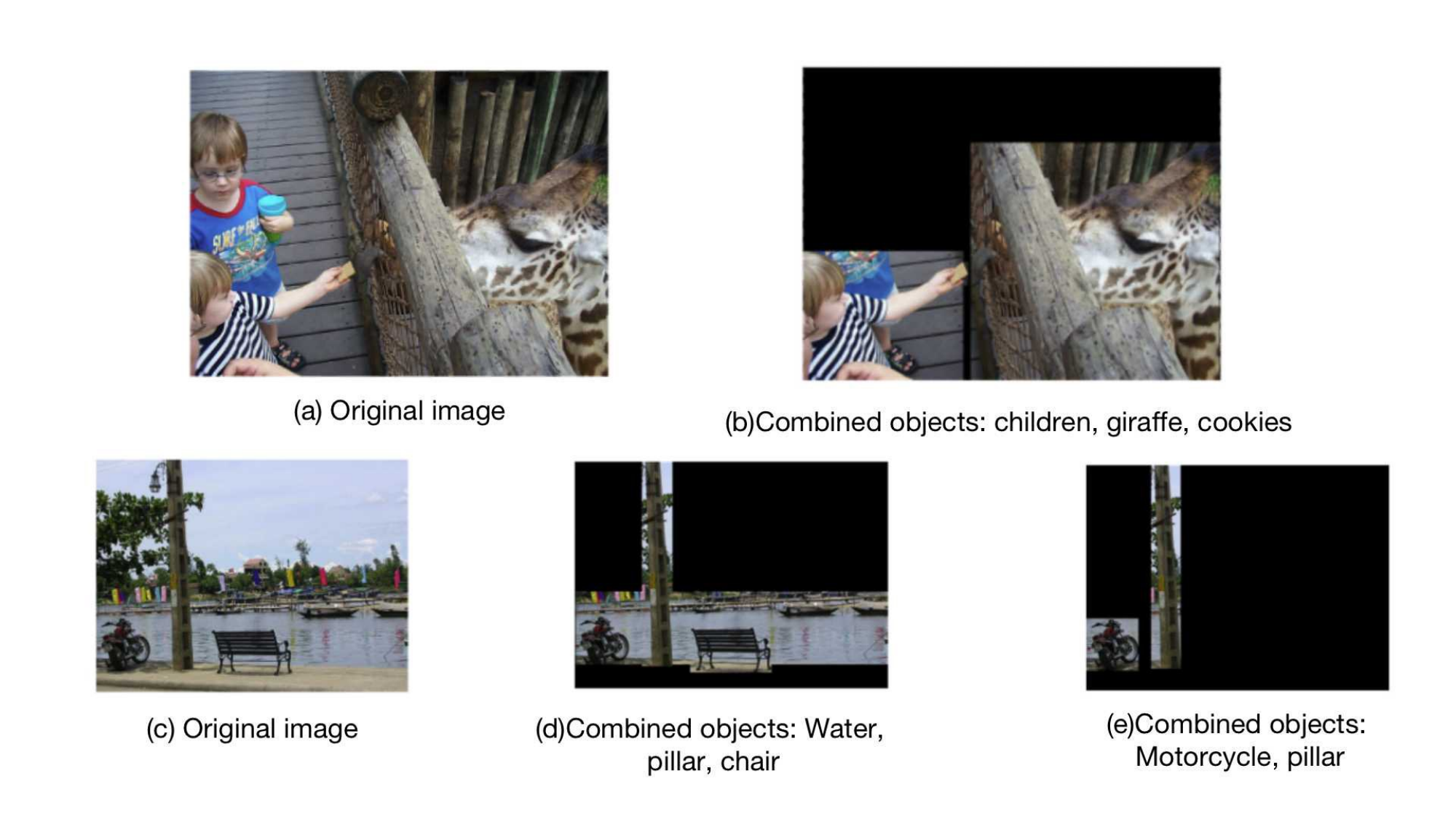}
    \caption{Example of two combined object sub-image}
    \label{fig:my_label}
\end{figure}
As shown in Figure 3, Figure 3(a) presents the original image, while Figure 3(b) showcases the corresponding combined object sub-image. It can be seen that the combined object sub-image retains information about the objects themselves and their relative positional information while removing other redundant details from the image, thus emphasizing the information that "a child is feeding cookies to a giraffe." Similar roles can be observed in Figures 3(c), 3(d), and 3(e).

\subsubsection{Features Extraction}

After constructing the combined object sub-image $I^c$, we used the pre-trained ResNet network \cite{he2016deep} to extract visual features as combined object features $v^c$ from the combined object sub-image.
\begin{equation}
    v^c=ResNet(I^c)
\end{equation}
where the combined object visual features $v^c \in R^{H'*K'}$, $H'$ is the dimension of the visual features, and $K'$ is determined by the size of the image and the ResNet.

\subsection{Language Modules}

Our language module adopts a sentence-by-sentence generation approach. It utilizes the visual features $v_t^c$ of the t-th composite object sub-image $I_t^c$ and the previously generated $t-1$ sentences ${y_1,y_2,...,y_{t-1}}$ to generate the next $t$-th sentence $y_t$. Our language model employs the VIRTEX method\cite{desai2021virtex} and the SCST (w/rep.penalty) method\cite{melas2018training}.

\subsubsection{VIRTEX Method}

The method based on VIRTEX utilizes a Transformer decoder as its language model. The input to the decoder consists of two parts: (1) the combined object visual features corresponding to the current $t$-th sentence, extracted through the ResNet network, and (2) the previously generated $t-1$ sentences. The output of the decoder is the current $t$-th sentence.

\subsubsection{SCST(w/rep. penalty) Method}

The language module in the method based on SCST(w/rep. penalty) is composed of a top-down attention LSTM and a language LSTM. The top-down attention LSTM is used to assess the importance of each feature region, while the language LSTM decodes the image features into textual descriptions. The input for the top-down attention LSTM consists of three parts: (1)the previous output of the Language LSTM, (2)the average pooled object features, and (3)the encoding of the previously generated word. The input for the Language LSTM includes the output from the top-down attention LSTM, and the object features weighted through an attention mechanism. The language LSTM outputs the corresponding description of the image.

Contrary to the original model, which uses visual features based on all objects detected, our approach uses the combined object visual features extracted through the pre-trained ResNet. This adjustment allows our language model to filter out extraneous information, leading to a more compelling image description generation.

\subsection{Aligning Objects and Sentences}

For the implementation of our approach, the crucial prerequisite is the alignment of objects and sentences in the dataset. However, no public fine-grained image description generation dataset exhibits this alignment property between objects and sentences. To address this issue, we decide to align objects and sentences automatically.

Given that the labels of object detection and objects in sentences are not always identical but semantically similar, we cannot rely solely on whether a sentence contains the object label to determine relevance. To tackle this, we employ word embedding technology to compute the semantic similarity between the object label and sentence, thus aligning the object with the sentence.

BERT\cite{devlin2018bert} is one of the most popular pre-trained language models, so we opted to use a pre-trained BERT to calculate the cosine similarity between the embedding of the object categories $E^O={E_1^O,E_2^O,...,E^O_K}$ and the word embedding in the sentence $E^w={E_1^w,E_2^w,...,E_N^w}$. With this semantic similarity, we can align the objects with the sentences.

\section{Experiments}

\subsection{Dataset and Baseline Model}

In these experiments, we utilize the Stanford Image Paragraph Captioning dataset\cite{krause2017hierarchical}. This dataset, sourced from MS COCO\cite{lin2014microsoft} and the Visual Genome\cite{krishna2017visual}, is among the most popular for fine-grained image description generation research. It contains 19,551 images, each with at least one paragraph of description.

We selected several popular fine-grained image description generation models for benchmarking to facilitate comparison: CAVP\cite{liu2018context}, SCST w/rep.penalty\cite{melas2018training}, IMAP\cite{xu2020interactive}, IMG+LNG\cite{ilinykh2020image}, VIRTEX\cite{desai2021virtex}, Regions-Hierarchical\cite{krause2017hierarchical}, and VTCM-Transformer\cite{guo2022matching}. These models have demonstrated strong competitiveness in fine-grained image description generation, with some even achieving state-of-the-art results.

\subsection{Experiment Setup}

In our experiments, Faster R-CNN employs a pre-trained model with the extracted object visual features having a dimension of 2048. 
In the method based on VIRTEX, both ResNet and Transformer decoders utilize pre-trained VIRTEX models without freezing the ResNet parameters. ResNet-50 is chosen for ResNet, with an output dimension of 2048. The parameters for the Transformer decoder are shown in the table. The training process involves a bidirectional Transformer decoder, whereas the generation process is unidirectional.

\begin{table}[]
\centering
\caption{Main parameters of the Transformer in VIRTEX}
\begin{tabular}{|l|l|}
\hline
Parameters                & Size \\ \hline
Dimension of hidden layer & 2048 \\ \hline
Number of layers          & 1    \\ \hline
Maximum length            & 150  \\ \hline
\end{tabular}
\end{table}

For the method based on SCST (w/ rep. penalty), ResNet utilizes a pre-trained ResNet-50 with frozen model parameters. The parameters for LSTM are detailed in the table.

\begin{table}[]
\centering
\caption{Main parameters of LSTM}
\begin{tabular}{|l|l|}
\hline
Parameters                & Size \\ \hline
Dimension of hidden layer & 1300 \\ \hline
Number of layers          & 2    \\ \hline
\end{tabular}
\end{table}

\subsection{Evaluation Metrics}

In addition to commonly used evaluation metrics in the field of fine-grained image description generation, such as BLEU\cite{papineni2002bleu}, CIDEr\cite{vedantam2015cider}, and METEOR\cite{denkowski2014meteor}, we propose new metrics based on the number of objects to more directly evaluate the performance of different methods in addressing issues of description repetition and omission. Expressly, we set four evaluation metrics based on the number of objects, which are:

(1)The number of described objects $|O_G|$: The average number of objects described per image in image description generation. A higher value indicates a more comprehensive image description, as it covers more objects.

(2)The number of objects Consistent with ground truth $|O_{G-Cap}|$: the average number of objects in a description that matches those in the ground truth. A higher value indicates the ability to describe more objects consistent with the ground truth.

(3)The coverage rate of described object $RC_{Cap}$ :the ratio of $|O_{G-Cap}|$ to $|O_G|$. A higher value indicates a better ability to describe objects consistent with the ground truth.
\begin{equation}
   RC_{Cap}=\frac{|O_{G-Cap}|}{|O_{Cap}|}
\end{equation}

(4)$Rep-4$ : the number of objects in the fine-grained image description that appears four or more times (inclusive). A lower score in this metric is preferable, as repeating an object four times or more is usually considered excessive repetition.

The first three metrics assess the descriptions' comprehensiveness, while $Rep-4$ gauges the repetition of descriptions.

\subsection{Experimental results and discussions}

\begin{table}
\caption{Comparison with other methods on Image Paragraph Captioning dataset. The best results for each evaluation metric are marked in bold.}
\resizebox{\linewidth}{!}{
\begin{tabular}{|l|l|l|l|l|l|l|}
\hline
Methods                           & METEOR         & CIDEr          & BLEU-1         & BLEU-2         & BLEU-3         & BLEU-4         \\ \hline
IMG + LNG                         & 11.28          & 26.04          & 24.96          & 13.82          & 8.04           & 4.6            \\
IMAP                              & \textbf{17.13} & 22.98          & \textbf{44.02} & 27.29          & 16.75          & 9.79           \\
Regions-Hierarchical              & 15.95          & 13.52          & 41.9           & 24.11          & 14.23          & 8.69           \\
CVAP                              & 16.83          & 21.12          & 42.01          & 25.86          & 15.33          & 9.26           \\
VTCM-Transformer                  & 16.88          & 26.15          & 40.93          & 25.51          & 15.94          & 9.96           \\ 
\hline
VIRTEX                            & 14.56          & 15.9           & 33.24          & 19.68          & 12.12          & 7.44           \\
VIRTEX + FIDG-JO(ours)               & 14.91          & 23.19          & 39.48          & 23.04          & 13.92          & 8.52           \\  \hline
SCST (w/ rep.penalty)             & 16.28          & 25.34          & 42.11          & 29.51          & 17.36          & \textbf{10.56} \\
SCST(w/ rep.penalty) + FIDG-JO(ours) & 16.73          & \textbf{28.46} & 42.35          & \textbf{29.57} & \textbf{18.21} & 10.54  \\ \hline
\end{tabular}
}
\end{table}

Table 3 compares VIRTEX and SCST (w/rep.penalty), incorporating the FIDG-JO method against other models. With the introduction of FIDG-JO, VIRTEX has improved on all metrics, especially CIDEr, which has seen a significant rise from 15.90 to 23.19. CIDEr is an evaluation metric designed explicitly for image description generation tasks, and it is considered the most crucial metric for assessing the performance of fine-grained image description generation. After incorporating FIDG-JO, SCST (w/rep.penalty) has also seen improvements in most metrics, particularly in CIDEr, which increased from 25.73 to 29.42.

The experimental results show that FIDG-JO significantly improves the CIDEr metric, which suggests that, compared to other methods, the descriptions generated by FIDG-JO contain more keywords, pay attention to more information in the image, and provide a more comprehensive description.

\begin{table}[]
\centering
\caption{Experimental results based on object-based evaluation metrics}\label{tab2}
\begin{tabular}{|l|l|l|l|l|}
\hline
Methods and ground truth                                 & $|O_G| $          & $|O_{G-Cap}|$ & $RC_{cap}$   & $Rep-4$         \\ \hline
Ground truth                                  & 6.38          & -              & -             & -             \\ \hline
VIRTEX                               & 4.93          & 3.82           & 59.9          & 0.39          \\
VIRTEX + FIDG-JO (ours)                 & 7.31          & 5.22           & 81.8          & 0.18          \\ \hline
SCST ( w/ rep.penalty)               & 6.05          & 5.34           & 83.7          & 0.25          \\
SCST ( w/ rep.penalty) + FIDG-JO (ours) & \textbf{7.62} & \textbf{5.63}  & \textbf{88.2} & \textbf{0.13} \\ \hline
\end{tabular}
\end{table}

Table 4 showcases the comparison of object-based evaluation metrics for VIRTEX and SCST (w/rep.penalty) before and after the introduction of FIDG-JO. It is noticeable that both VIRTEX and SCST (w/rep.penalty) have shown significant improvements in various metrics concerning description comprehensiveness after incorporating FIDG-JO, which indicates that the FIDG-JO can describe more objects and has a higher coverage rate for ground truth objects. At the same time, the metric of repeating the same object four times or more has noticeably decreased which fully illustrates that the FIDG-JO can effectively mitigate the problems of description repetition and omission.

\subsection{Experimental examples and analyses}

Figure 4 presents examples of image descriptions from VIRTEX, SCST (w/ rep. penalty), and after incorporating FIDG-JO.

\begin{figure}
    \centering
    \includegraphics[width=1.0\textwidth]{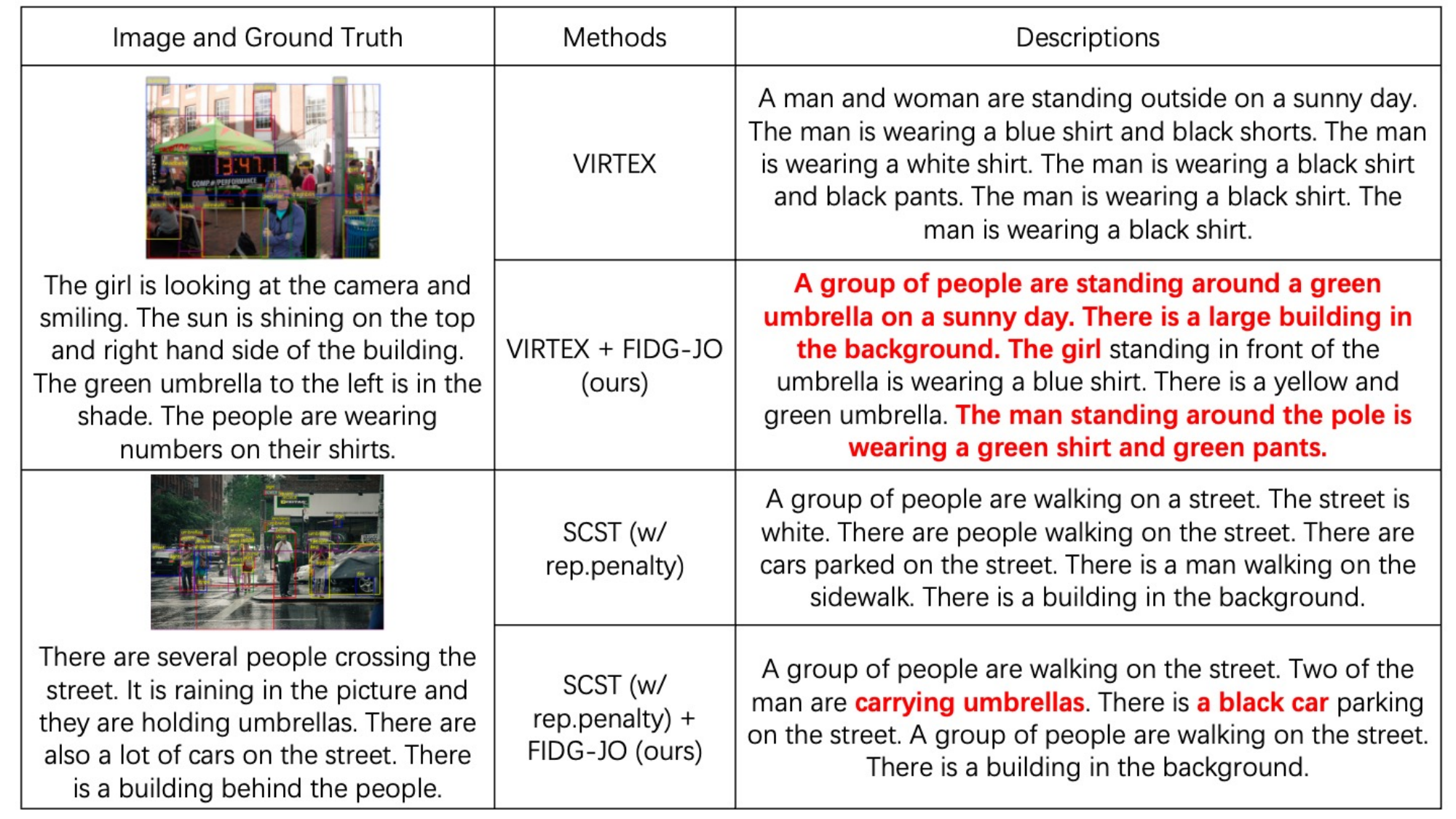}
    \caption{Examples of image descriptions generated by VIRTEX, SCST (w/ rep. penalty), and after incorporating FIDG-JO}
    \label{fig:my_label}
\end{figure}

In the descriptions provided by VIRTEX, the number of objects consistent with the annotations is four, while for VIRTEX+FIDG-JO, the number is six, indicating that FIDG-JO can effectively enhance the coverage of annotated objects. VIRTEX tends to repeat objects at a higher rate, repeating the main subject (man) five times, and the shirt appears four times. With VIRTEX+FIDG-JO, the repetition is reduced, with the umbrella appearing three times and the shirt appearing two times. Analyzing the descriptions, we find that the occurrence of objects up to three times is relatively common and aligns with the logical description. In contrast, more frequent occurrences are abnormal and considered repetitions, which shows that FIDG-JO can significantly reduce the repetition in descriptions. VIRTEX frequently describes the subject as a man, while the image's subject should be female; VIRTEX+FIDG-JO accurately describes the subject as a girl. Also, VIRTEX has a lower accuracy rate for color recognition, while VIRTEX+FIDG-JO can recognize colors more accurately, indicating that FIDG-JO improves the accuracy of descriptions.

For SCST (w/ rep. penalty), the number of objects consistent with the annotations is four. With SCST (w/ rep. penalty) + FIDG-JO, it is five, suggesting that FIDG-JO can increase the annotation coverage. SCST (w/ rep. penalty) includes one object inconsistent with the annotation, while SCST (w/ rep. penalty) + FIDG-JO is entirely consistent with the annotation. SCST (w/ rep. penalty) ignores the rain in the image, but SCST (w/ rep. penalty) + FIDG-JO's description includes an umbrella, signifying that FIDG-JO can make the description more comprehensive. In SCST (w/ rep. penalty)'s description, the word 'street' appears four times, creating an excessive repetition. With SCST (w/ rep. penalty) + FIDG-JO, no descriptions exceed three repetitions, showing that FIDG-JO can reduce the redundancy in descriptions.

These experimental examples further confirm the effectiveness of FIDG-JO in solving the problems of repetitive and omitted descriptions.

\section{Conclusion}

To address the two primary challenges in fine-grained image description generation — description repetition and omission, we propose a Fine-grained Image Description Generation method based on Joint Objectives (FIDG-JO). FIDG-JO ingeniously integrates the advantages of image-level and object-level visual features, accurately capturing object information while excluding irrelevant image information. Additionally, we introduce an object penalty mechanism to limit the number of times an object is redundantly described. Experiments validate the effectiveness of FIDG-JO in mitigating description repetition and omission problems. Furthermore, since existing evaluation metrics cannot intuitively reflect the issues of description repetition and omission in descriptions, we propose object-based evaluation metrics. By examining the occurrence of objects, we can more intuitively reflect the method's performance.

There are still parts of our method that need to be perfected. Manual alignment of objects and sentences would consume substantial human resources, so we decided to automate alignment using word embedding. However, many object-sentence alignment results might need to be revised. Thus, we will explore more efficient alignment methods to improve object-sentence alignment accuracy in future research.

\end{document}